%% file: main.tex
\documentclass[conference]{IEEEtran}
\input{preamble}

\hypersetup{
  pdftitle={Low-Density 3D Point Cloud Classification},
  pdfauthor={Ahmed Baha Ben Jmaa and Faten Chaieb},
  pdfkeywords={low-density 3D point cloud, point cloud classification, PointNet++}
}

\title{Low-Density 3D Point Cloud Classification}

\author{%
\IEEEauthorblockN{Ahmed Baha Ben Jmaa}
\IEEEauthorblockA{\textit{Efrei Research Lab} \\
\textit{Paris Panthéon-Assas University}, France\\
ahmedbaha.benjmaa@outlook.fr}
\and
\IEEEauthorblockN{Faten Chaieb}
\IEEEauthorblockA{\textit{Efrei Research Lab} \\
\textit{Paris Panthéon-Assas University}, France\\
faten.chakchouk@efrei.fr}
}

\newcommand{\AddFirstPagePublicationNotice}{%
  \AddToShipoutPictureFG*{%
    \AtPageLowerLeft{%
      \raisebox{0.14in}{%
        \hspace{0.65in}%
        \parbox{\dimexpr\paperwidth-1.30in\relax}{%
          \fontsize{5.8}{6.4}\selectfont\raggedright
          \textcopyright~2024 IEEE. Personal use of this material is permitted. Permission from IEEE must be obtained for all other uses, in any current or future media, including reprinting/republishing this material for advertising or promotional purposes, creating new collective works, for resale or redistribution to servers or lists, or reuse of any copyrighted component of this work in other works.\\[-0.1ex]
          This is the authors' accepted manuscript of the paper published in the \emph{2024 IEEE 36th International Conference on Tools with Artificial Intelligence (ICTAI)}, pp.~617--623. Version of record: \href{https://doi.org/10.1109/ICTAI62512.2024.00093}{doi:10.1109/ICTAI62512.2024.00093}.%
        }%
      }%
    }%
  }%
}

\begin{document}
\AddFirstPagePublicationNotice
\maketitle

\input{sections/00_abstract}
\input{sections/01_introduction}
\input{sections/02_benchmarking_study}
\input{sections/03_ld_pointnet}

\input{sections/04_experimental_evaluation}

\input{sections/05_conclusion}

\bibliographystyle{IEEEtran}
\bibliography{references}

\end{document}

%% file: preamble.tex
\usepackage{amsmath,amssymb,amsfonts}

\usepackage{graphicx}
\graphicspath{{figures/}}
\usepackage{xcolor}
\usepackage{array}
\usepackage{booktabs}
\usepackage{colortbl}
\usepackage{multirow}
\usepackage{siunitx}
\usepackage{subfig}

\usepackage{cite}
\usepackage{balance}

\usepackage{pifont}
\newcommand{\cmark}{\ding{51}}
\newcommand{\xmark}{\ding{55}}

\definecolor{LightGray}{gray}{0.9}
\newcommand{\parasection}[1]{\vspace{.05in}\noindent\textbf{#1}}

\usepackage{eso-pic}

\usepackage[hidelinks]{hyperref}

%% file: sections/00_abstract.tex
\begin{abstract}
3D point cloud classification, a task in computer vision, has recently gained substantial interest due to its extensive applications across various domains such as augmented and virtual reality, robotics, and autonomous driving.
However, classifying 3D point clouds becomes challenging in real-world scenarios where the density levels vary. These variations are due to limitations inherent to current sensing technologies and variable environmental conditions, which often prevent the capture of detailed and comprehensive geometric representations. Existing classification methods, which have demonstrated impressive results on dense point clouds, may be affected and could experience reduced performance when dealing with low-density point clouds. Addressing this challenge, this paper introduces LD-PointNet++, an architecture optimized for the classification of low-density point clouds. To pave the way for LD-PointNet++, we conduct the first benchmarking study that systematically evaluates state-of-the-art point cloud classification methods across varying densities, revealing significant performance declines in low-density environments. 
Experimental results on the ModelNet benchmark dataset demonstrate that LD-PointNet++ not only outperforms these methods in low-density scenarios but also demonstrates consistent accuracy above 90\% across different density levels. 
\end{abstract}


%% file: sections/01_introduction.tex
\section{Introduction}
Nowadays, point cloud, a geometric representation of three-dimensional (3D) data, is gaining increasing interest in emerging computer vision applications~\cite{9127813,9857670}. It is finding more and more uses in domains including augmented and virtual reality, robotics, and autonomous driving, thanks to rapid advancements in acquisition techniques such as laser scanners and depth cameras~\cite{9857670}.
A 3D point cloud represents the direct result of capturing the spatial distribution of objects and offers the advantage of being both simple and precise. It provides rich geometric information, allowing for a faithful reproduction of the real world.

However, real-world scenarios often present challenges in acquiring point clouds, primarily due to variable environmental conditions and sensor limitations. These factors frequently result in low-density point clouds, characterized by sparse points that occur due to increased distances between the sensor and the 3D object/scene, or due to the constraints of the capture devices. These limitations compromise the ability to capture fine details of 3D objects/scenes, which are essential for computer vision tasks. For example, emerging applications like autonomous driving, which require efficient processing methods for point clouds, encounter a reduction in point density as the spatial separation from the sensor to the object extends.

Consequently, classifying 3D point clouds with low density poses several challenges. Indeed, a limited number of points in a point cloud can lead to an incomplete or even incomprehensible geometric representation of the 3D object/scene, resulting in decreased classification accuracy. Although state-of-the-art methods for point cloud classification have shown impressive results in scenarios where point clouds are very dense, they fail to maintain comparable performance in low-density scenarios.
In response to this limitation, this paper introduces LD-PointNet++, a model specifically designed for low-density point cloud classification. The main objective is to address the challenges posed by variations in point cloud density, ensuring consistent and accurate classification outcomes across different scenarios.

\parasection{Contributions.}
Our key contributions are summarized as follows:
\textbf{(1)} We present the first benchmarking study of state-of-the-art methods for low-density 3D point cloud classification through a robustness study on input density. This study reveals that the performance of these methods significantly degrades in low-density scenarios, providing crucial insights into their limitations.
\textbf{(2)} We propose \textit{LD-PointNet++}, a novel variant specifically optimized for low-density 3D point cloud classification. LD-PointNet++ not only consistently outperforms state-of-the-art methods in low-density scenarios but also maintains an accuracy above 90\% across varying densities. We demonstrate that no previous method achieves such high accuracy with low density.

\parasection{Paper Organization.} 
In the following, Section~\ref{section:RobustnessAnalysis} presents a benchmarking study of 3D point cloud classification methods, focusing on their robustness to density variations through experimental evaluations.
Section~\ref{section:LD-PointNet++} introduces LD-PointNet++, explaining its working principle in low-density scenarios.
Section~\ref{section:Experiments} focuses on evaluating the effectiveness of LD-PointNet++ in classifying low-density 3D point clouds.
Finally, Section~\ref{section:Conclusion} concludes the paper.

%% file: sections/02_benchmarking_study.tex
\section{Unveiling the Challenge: Low-Density Point Cloud Classification - Benchmarking Study}
\label{section:RobustnessAnalysis}
    In this section, we investigate the robustness of 3D point cloud classification methods to density variation. Real-world scenarios often involve incomplete point cloud acquisition due to factors such as acquisition conditions and sensor characteristics. Consequently, point density can vary across different regions of the 3D object/scene, which poses challenges for the analysis and processing of point clouds. Low-density 3D point clouds, in particular, present a challenge for classification models as they may contain insufficient data to accurately classify the 3D object/scene. Understanding how 3D point cloud classification methods perform under such conditions is crucial.
    
    \subsection{Methods Overview}
    In this section, we provide a concise overview of the working principle of point cloud classification methods. 
    
        \parasection{Pointwise MLP Methods.}
        PointNet~\cite{PointNet} is the pioneering model that processes point clouds directly as input to the neural network. It uses multiple Multilayer Perceptron (MLP) layers to learn feature for each point, and finally aggregates them into a global feature using a max pooling layer. PointNet learns each point independently, therefore it does not capture local structure and interactions between points. To address these limitations, PointNet++~\cite{PointNet++} is introduced as a hierarchical network that captures local features for each point from its neighborhood. The idea is to learns features by considering the local context of each point. To achieve this, PointNet++ defines a number of abstraction levels. Each abstraction level includes three distinct layers: a sampling layer, a grouping layer, and a PointNet layer. At the last abstraction level, features are aggregated using max pooling to obtain a learnable representation. This feature vector is then passed through a fully connected classifier network to obtain the score for each class. Following PointNets, many point-wise MLP methods have been introduced, either based on or inspired by PointNet++~\cite{PointNeXt,PointMLP,9127813}.

        \parasection{Convolution-Based Methods.}
        These methods define convolutional kernels for point cloud. Several definitions of convolution operators have been proposed to adapt convolutional neural networks to point cloud~\cite{Pointwise,KPConv,Zhang2020AECNN,Zhang2022RIConv++,9127813}. 
        In general, for the majority of approaches, for a point cloud $\boldsymbol{\mathcal{P}}=\left\{p_{i} \in \mathbb{R}^{3}\right\}_{i<N} \in \mathbb{R}^{N \times 3}$, the input feature map in a convolutional layer, denoted as $\boldsymbol{\mathcal{F}}$, can be expressed as $\boldsymbol{\mathcal{F}}=\left\{f_{i} \in \mathbb{R}^{D}\right\}_{i<N} \in \mathbb{R}^{N \times D}$.
        The convolution of $\boldsymbol{\mathcal{F}}$ by a kernel $g$ at a point $p \in \mathbb{R}^3$ with neighborhood $\mathcal{N}_p$ is defined as:
        \begin{equation}
            (\mathcal{F} * g)(p)=\sum_{p_i \in \mathcal{N}_p} g\left(p_i-p\right) f_i.
            \label{eq:Convolution}
        \end{equation}
        In this definition, two parts are important: the definition of the neighborhood $\mathcal{N}_p$ and the definition of the kernel function~$g$. The two types of neighborhoods commonly used in the literature are the spherical neighborhood (defined by a radius) and the K-Nearest Neighbors (KNN) neighborhood~\cite{9127813}. 
        The kernel function $g$ is a crucial element in~\eqref{eq:Convolution}, differentiating state-of-the-art methods. This function takes as its input the positions of neighbors centered around the point~$p$ ($q_i = p_i - p$) and produces a weight matrix in $\mathbb{R}^O$, where $O$ represents the dimension of the output feature vector.
        
        \parasection{Graph-Based Methods.}
        Graph-based methods~\cite{dgcnn,Rgcnn,9127813} are generally composed of two parts: \textbf{(1)}~graph construction and \textbf{(2)}~feature learning and clustering. Each point in the point cloud is considered as a vertex (node). From the neighborhood of each point, edges are generated. Formally, to create a directed graph $G=(\mathcal{P}, \mathcal{E})$ from a point cloud $\mathcal{P}$, we connect each vertex $i$ to every vertex $j$ within its local spatial neighborhood.
        Learning is subsequently conducted in either spatial or spectral domains. In this way, these methods are grouped into two classes: Methods based on spatial structure analysis techniques and Methods based on spectral graph theory~\cite{9127813}.
        \begin{figure}[tbp]
            \centering
            \includegraphics[width=\columnwidth]{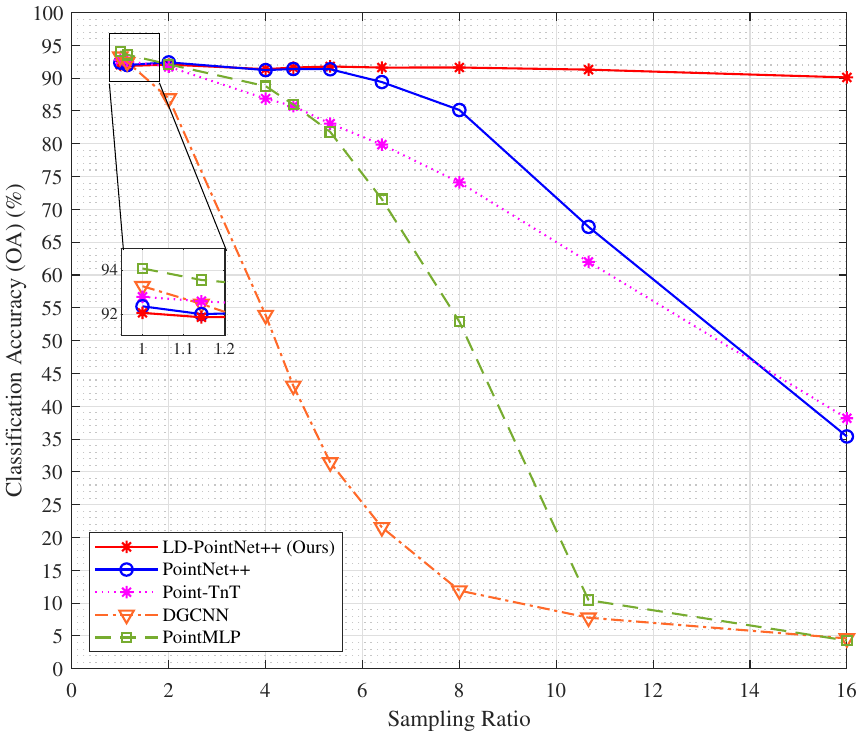}
            \caption{Robustness to sampling density variation.}
            \label{fig:RobustesseSampling}
        \end{figure}
         
    \subsection{Experimental Setup and Results}
    \begin{figure*}[tbp]
        \centering
        \resizebox{0.84\textwidth}{!}{
        \setlength{\tabcolsep}{3pt} 
        \renewcommand{\arraystretch}{1.5} 
        \begin{tabular}{c>{\centering\arraybackslash}m{0.22\linewidth}>{\centering\arraybackslash}m{0.22\linewidth}>{\centering\arraybackslash}m{0.22\linewidth}>{\centering\arraybackslash}m{0.22\linewidth}} 
            \textbf{Model} & \textbf{PointNet++}~\cite{PointNet++} & \textbf{Point-TnT}~\cite{Point-TnT} & \textbf{DGCNN}~\cite{dgcnn} & \textbf{PointMLP}~\cite{PointMLP} \\
            \textbf{ES1} & \includegraphics[width=\linewidth]{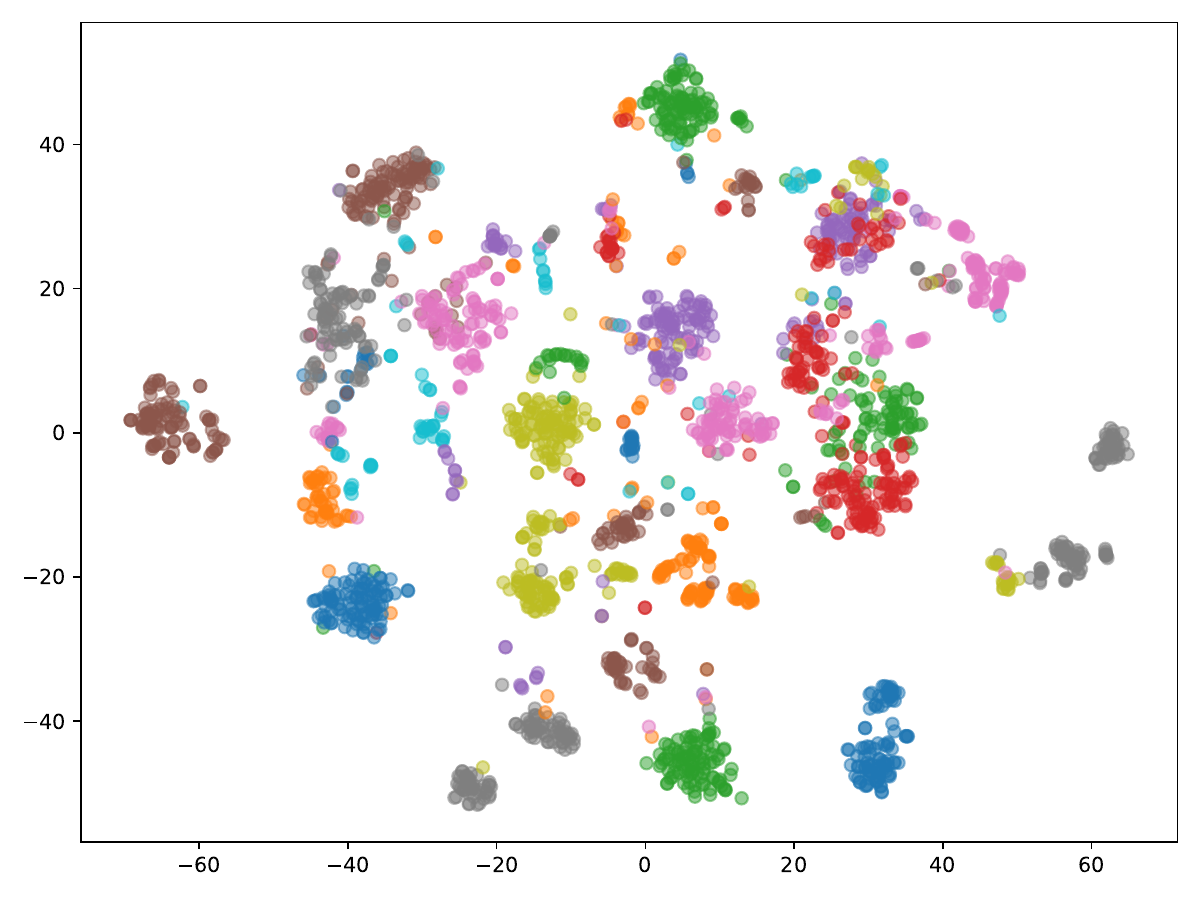} & \includegraphics[width=\linewidth]{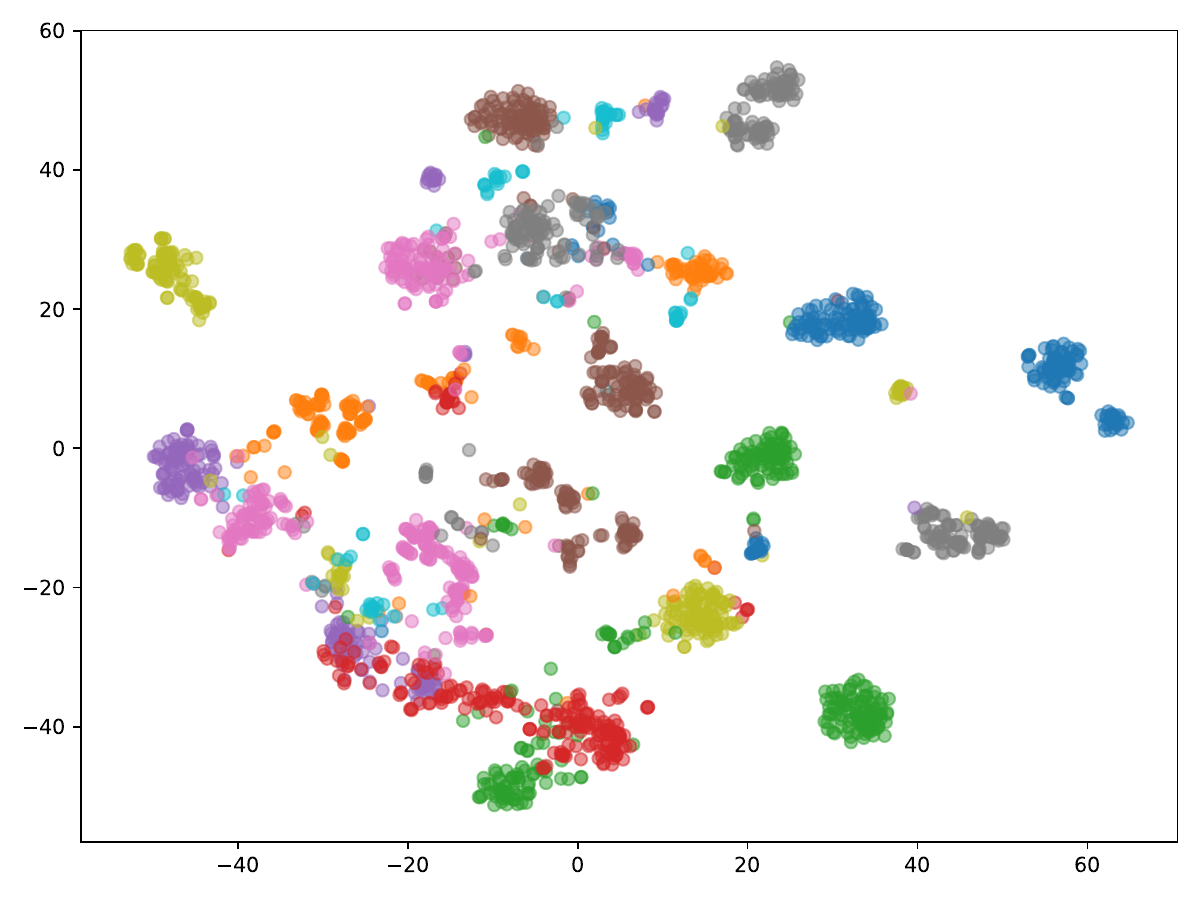} & \includegraphics[width=\linewidth]{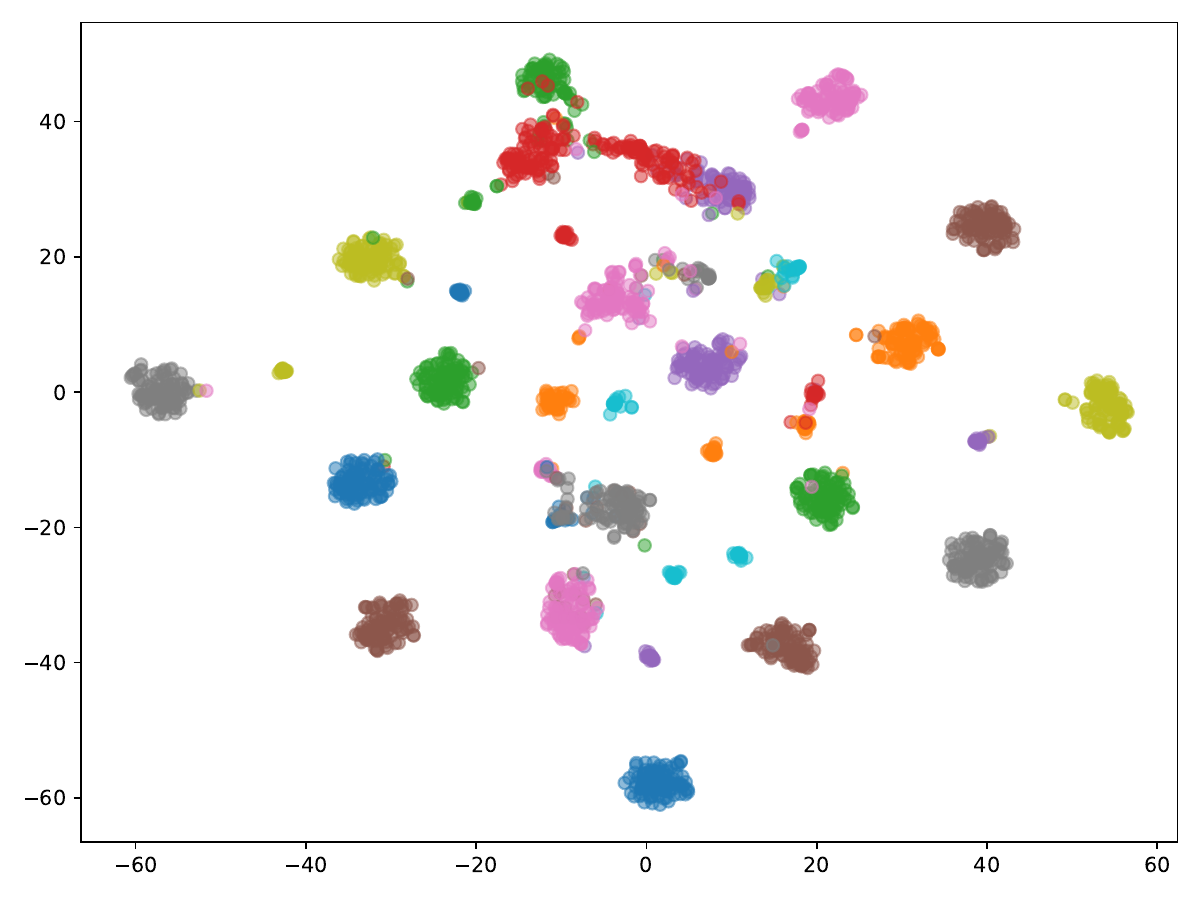} & \includegraphics[width=\linewidth]{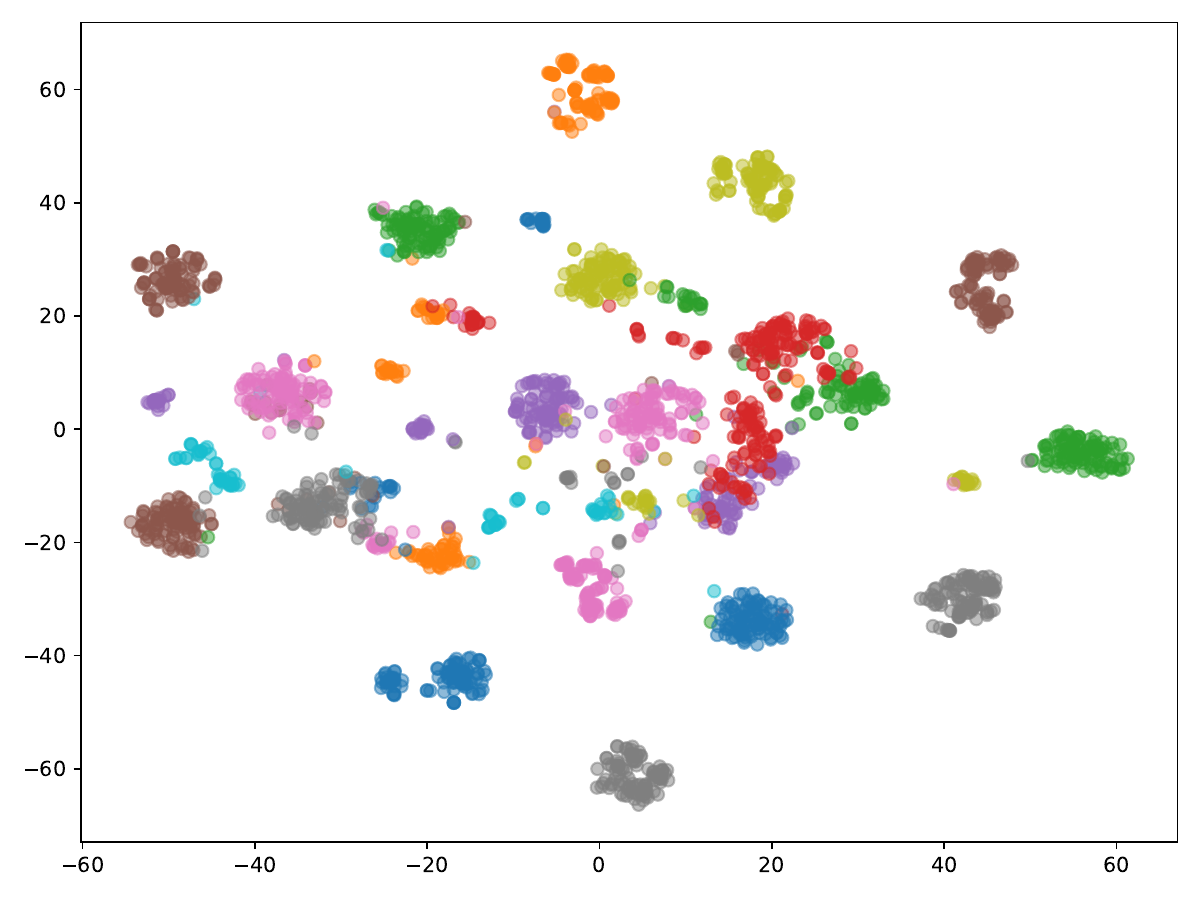}\\
            \textbf{ES2} & \includegraphics[width=\linewidth]{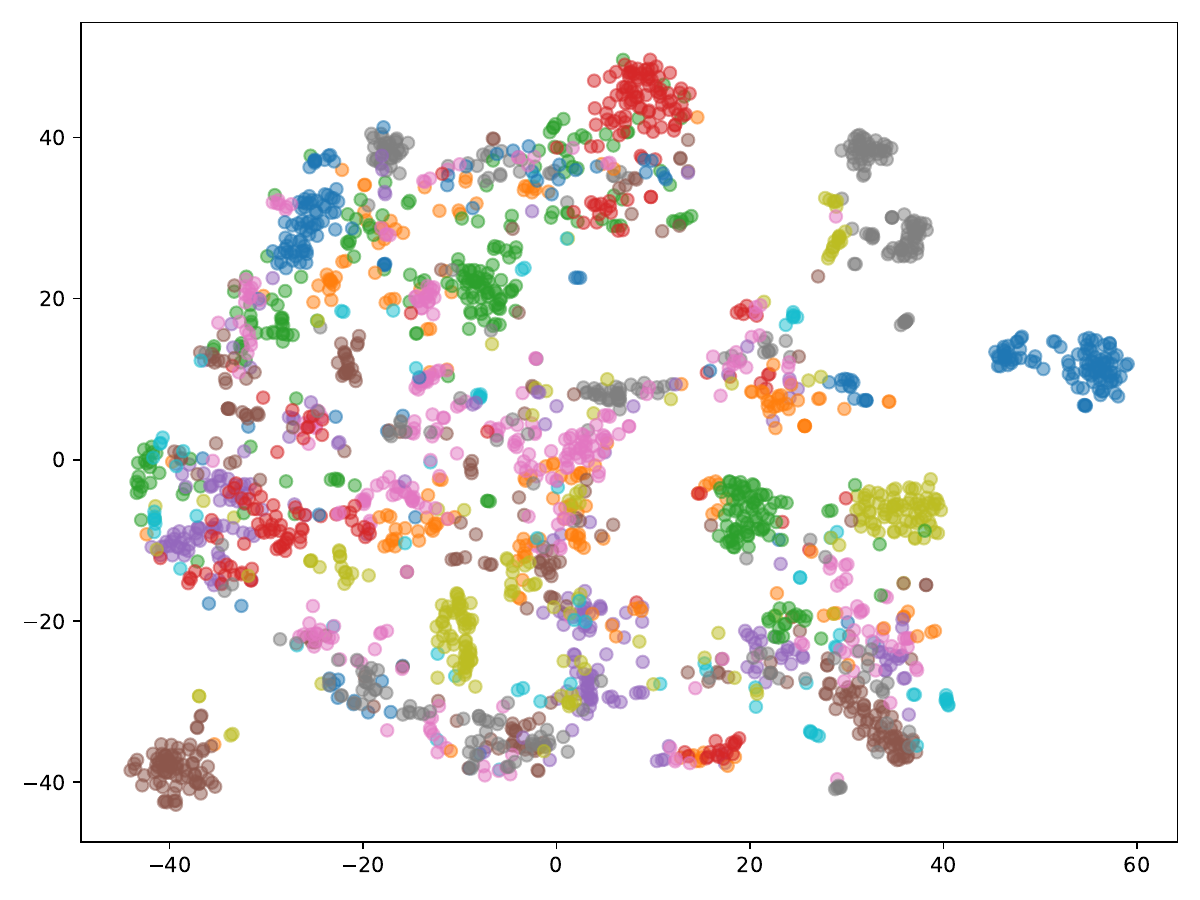} & \includegraphics[width=\linewidth]{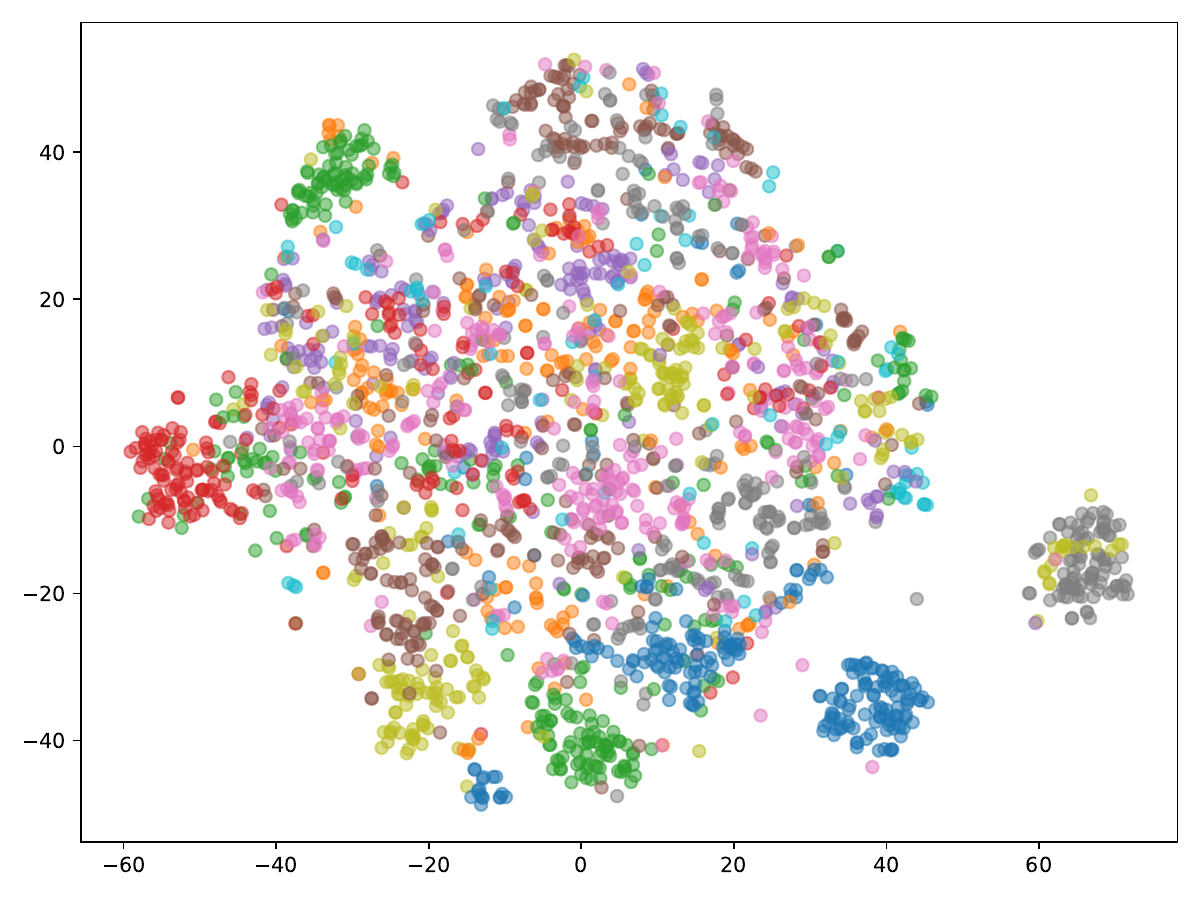} & \includegraphics[width=\linewidth]{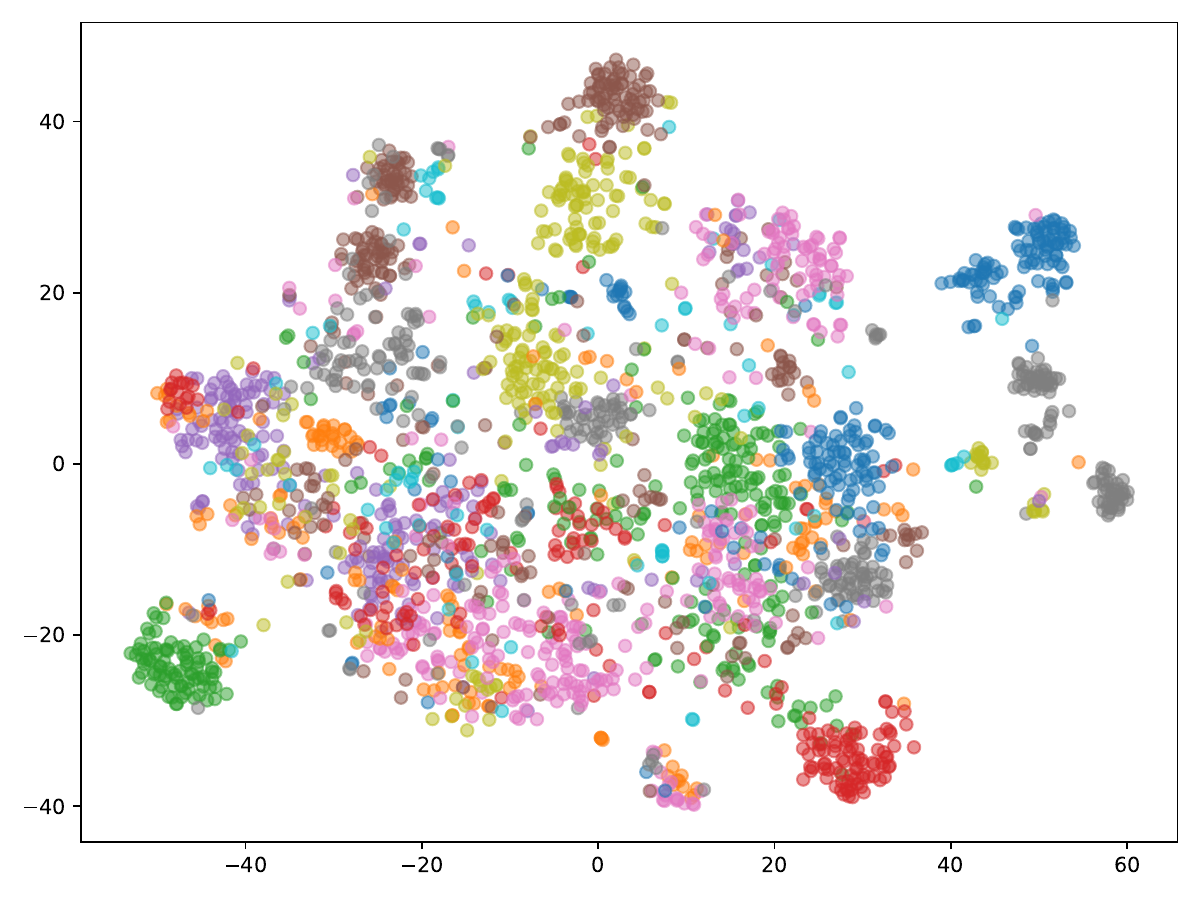} & \includegraphics[width=\linewidth]{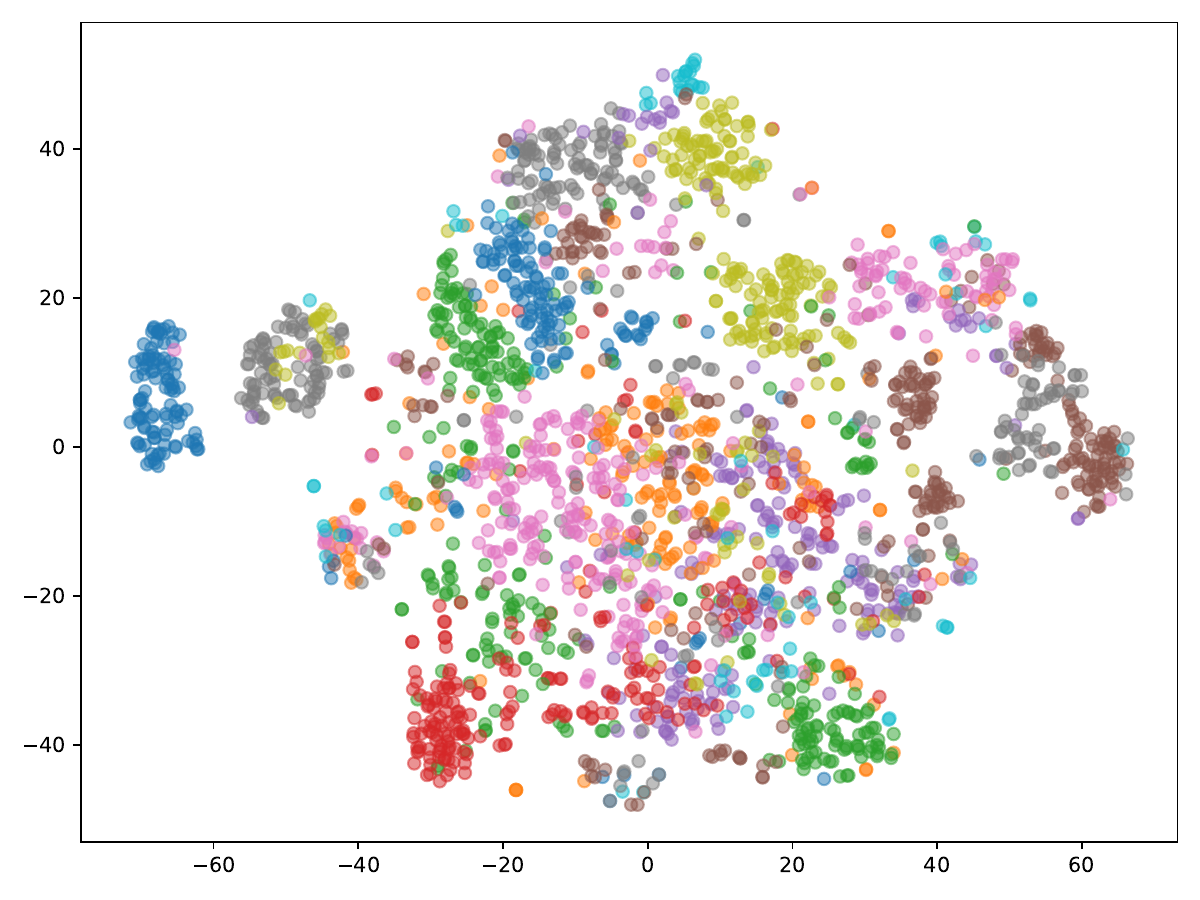}\\
        \end{tabular}
        }
        \caption{Comparison of t-SNE visualization for different models under experimental setups ES1 and ES2.}
        \label{fig:tsne-all-models}
    \end{figure*}
        In this section, we assess the robustness of existing 3D point cloud classification methods to variations in density. We focus on the most representative methods, including PointNet++~\cite{PointNet++}, Point-TnT (Point Transformer-in-Transformer)~\cite{Point-TnT}, DGCNN~\cite{dgcnn}, and PointMLP~\cite{PointMLP}. This study does not include other recent methods mentioned in~\cite{9127813} because they inherently require a large minimum number of points in the point cloud to function. 
        To conduct the evaluation, we train and evaluate candidate methods using the ModelNet40 variant of the ModelNet dataset~\cite{ModelNet}, which is widely adopted as a benchmark for point cloud classification. This dataset comprises 12,311~CAD models categorized into 40 object classes. The standard data split is followed, resulting in a training set of 9,843 objects and a test set containing 2,468 objects. Each input point cloud consists of 1024 points, with each point represented by attributes~$(x, y, z)$ denoting 3D Euclidean coordinates. 
        The evaluation involves training these methods on the complete dataset and subsequently testing them on subsets of points with varying densities. 
        This methodology—training on dense point clouds and testing on sparse ones—is designed to reflect real-world scenarios, where dense data might be available during the training phase but not during application. This approach is well-supported in the literature and aligns with practical applications as outlined in~\cite{PointNet++}.
        Density variations are achieved by randomly sampling the dataset, as proposed in~\cite{PointNet++}, while preserving the original distribution pattern of the point cloud. 
        
        In Fig.~\ref{fig:RobustesseSampling}, we present the test accuracy as a function of the sampling ratio. The sampling ratio is defined as~$\frac{N}{k} = \frac{1024}{k}$, where~$N$ is the number of original points, and $k$ represents the number of sampled points~($k \in \{64, 96, 128, 160, 192, 224, 256, 512, 1024\}$).
        
        To assess the resilience of these methods for low-density 3D~point cloud classification, we conducted experiments using a sampling ratio corresponding to 64 points per point cloud during the test phase, thereby simulating low-density scenarios. Subsequently, we focus on the following two Experimental Setups~(ES): \textbf{ES 1} - Training and testing using the entire point cloud~(1024 points). \textbf{ES 2} - Training using the entire point cloud (1024 points) and testing using only a few points~(64 points).
        For the evaluation of model performance, Table~\ref{table:Performances_Models} presents the following classification metrics: accuracy, precision, recall, and F1-score, computed on the test set across all experimental configurations.
        Finally, we apply t-SNE (t-distributed Stochastic Neighbor Embedding)~\cite{t-SNE}~to the learned features of the different models under the two experimental setups. The results are summarized in Fig.~\ref{fig:tsne-all-models}.
        \begin{table}[!htb]
            \caption{Performance of 3D point cloud classification models in two setups: full training and testing on 1024 points (ES1), and training on 1024 points with testing on 64 points~(ES2).}
            \label{table:Performances_Models}
            \sisetup{table-format=-1.2}
            \resizebox{\columnwidth}{!}{
                \begin{tabular}{@{}llcc@{}}
                    \toprule
                    \multirow{2}{*}[-0.5ex]{\textbf{Method}} & \multirow{2}{*}[-0.5ex]{\textbf{Metrics}} & \multicolumn{2}{c}{\textbf{Experimental Setup}} \\
                    \cmidrule(lr){3-4}
                    & & \textbf{ES1} & \textbf{ES2} \\
                    \midrule
                    \multirow{4}{*}{PointNet++~\cite{PointNet++}} 
                        & Accuracy  & 92.42\% & 35.45\% (\textcolor{red}{$\downarrow$ 56.97\%}) \\
                        & Precision & 92.55\% & 61.51\% (\textcolor{red}{$\downarrow$ 31.04\%}) \\
                        & Recall    & 92.42\% & 35.45\% (\textcolor{red}{$\downarrow$ 56.97\%}) \\
                        & F1-score  & 92.44\% & 34.66\% (\textcolor{red}{$\downarrow$ 57.78\%}) \\
                    \midrule
                    \multirow{4}{*}{Point-TnT~\cite{Point-TnT}} 
                        & Accuracy  & 92.79\% & 38.17\% (\textcolor{red}{$\downarrow$ 54.62\%}) \\
                        & Precision & 92.79\% & 62.04\% (\textcolor{red}{$\downarrow$ 30.75\%}) \\
                        & Recall    & 92.79\% & 38.17\% (\textcolor{red}{$\downarrow$ 54.62\%}) \\
                        & F1-score  & 92.68\% & 35.26\% (\textcolor{red}{$\downarrow$ 57.42\%}) \\
                    \midrule
                    \multirow{4}{*}{DGCNN~\cite{dgcnn}} 
                        & Accuracy  & 93.27\% & 4.62\% (\textcolor{red}{$\downarrow$ 88.65\%}) \\
                        & Precision & 93.59\% & 6.28\% (\textcolor{red}{$\downarrow$ 87.31\%}) \\
                        & Recall    & 93.27\% & 4.62\% (\textcolor{red}{$\downarrow$ 88.65\%}) \\
                        & F1-score  & 93.39\% & 2.35\% (\textcolor{red}{$\downarrow$ 91.04\%}) \\
                    \midrule
                    \multirow{4}{*}{PointMLP~\cite{PointMLP}} 
                        & Accuracy  & 94.08\% & 4.29\% (\textcolor{red}{$\downarrow$ 89.79\%}) \\
                        & Precision & 94.07\% & 10.37\% (\textcolor{red}{$\downarrow$ 83.70\%}) \\
                        & Recall    & 94.08\% & 4.29\% (\textcolor{red}{$\downarrow$ 89.79\%}) \\
                        & F1-score  & 94.04\% & 4.51\% (\textcolor{red}{$\downarrow$ 89.50\%}) \\
                    \bottomrule
                \end{tabular}
            }
        \end{table}
        \begin{figure*}[tbp]
            \centering
            \includegraphics[width=0.75\textwidth]{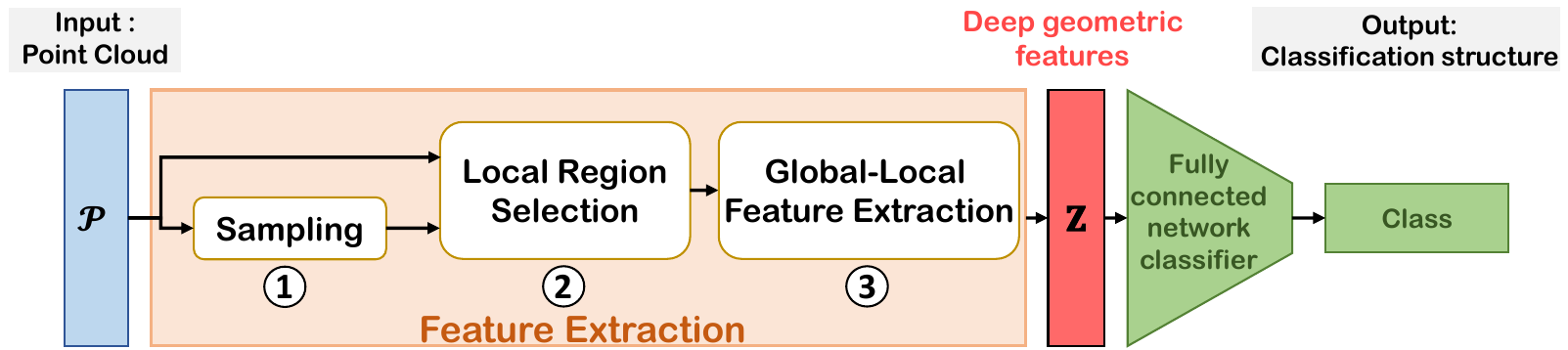}
            \caption{Illustration of LD-PointNet++ architecture.}
            \label{fig:LD-PointNet++}
        \end{figure*}
    \subsection{Findings and Analysis}
    \parasection{How robust are state-of-the-art point cloud classification models to variations in input density?}
    The results presented in Fig.~\ref{fig:RobustesseSampling} demonstrate the sensitivity of the evaluated point cloud classification models to density variation. The performance of the models compared in this study dramatically degrades when dealing with low-density 3D point cloud, especially with less than 160 input points, corresponding to a sampling ratio of 6.4. This degradation in accuracy is huge and may have implications in real-world applications where point clouds can be low-density. For example, PointMLP, considered one of the top-performing models in dense point cloud classification, has the most significant sensitivity to input density variations. With a degradation margin of up to~89.79\%, PointMLP's performance degrades significantly in low-density scenarios. On the other hand, PointNet++ shows some resilience compared to other models at different sampling rates. However, even PointNet++ struggles to achieve satisfactory results, achieving only 34.45\% accuracy when operating with 64~points per input point cloud.
    
    \parasection{How do state-of-the-art point cloud classification models perform in low-density 3D point cloud classification?}
    The robustness analysis focused on evaluating the models in two experimental configurations, ES1 and ES2~(see Table~\ref{table:Performances_Models}). The results obtained in ES1, which serve as a reference for evaluating the performance of dense point cloud classification, show that DGCNN and PointMLP achieve reasonably high performance on all metrics: DGCNN and PointMLP respectively reach an accuracy of 93.27\% and 94.08\%. However, the performance of these models drops dramatically when tested with only a limited number of points (64 points) while being trained on the entire point cloud (1024 points). This decrease in accuracy and other metrics indicates that the models fail to generalize in low-density scenarios. This supports the hypothesis that state-of-the-art point cloud classification methods, which excel in dense point cloud scenarios, face significant challenges in maintaining comparable performance in low-density scenarios. 
    The analysis of t-SNE visualizations of the features of different models under ES1 and ES2 further reinforces this observation (see Fig.~\ref{fig:tsne-all-models}). The good feature separation for DGCNN and PointMLP in the dense point cloud scenario (ES1) shows that these models effectively capture the data variability in dense point clouds. However, the poor feature separation observed for all four models in the low-density scenario (ES2) indicates their limitations in separating features in low-density point clouds. This finding aligns with the performance degradation observed in the quantitative evaluation, again confirming the difficulty of the models in handling low-density scenarios.
    
    This robustness study indicates that state-of-the-art models face significant challenges in maintaining their performance when dealing with low-density point clouds. Critical implications for real-world point cloud classification applications may arise. This emphasizes the need to improve the generalization capabilities of these models for more robust point cloud classification systems suitable for a wider range of applications.

%% file: sections/03_ld_pointnet.tex
\section{LD-PointNet++: A Proposed Method for Low-Density 3D Point Cloud Classification}
\label{section:LD-PointNet++}
    \subsection{Problem Formulation}
    Low-density 3D point cloud classification is a classification task that assigns a specific label to an object/scene represented by a point cloud with reduced density. The process of developing a deep learning model for this task involves training with dense 3D point clouds and then inferring on low-density point clouds during the testing phase. This methodology ensures the model's ability to generalize and accurately classify point clouds with varying densities. 
    Thus, the problem is formulated as follows. The input data during the training phase consists of a dense 3D~point cloud, denoted as $\mathcal{P}_{train} = \{p_1, \ldots, p_n\}$. During the testing phase, the model is presented with a low-density 3D point cloud, simulating a less accurate acquisition process, denoted as $\mathcal{P}_{test} = \{p_1, \ldots, p_m\}$, where $m < n$. The main goal is to design a classification function, denoted as~$f_c$, capable of classifying low-density point clouds. $f_c$ must convert each input point cloud into a corresponding probability for each category, while demonstrating the ability to generalize effectively across variable data densities. 
    
    \subsection{Model Overview}
        LD-PointNet++, as illustrated in Fig.~\ref{fig:LD-PointNet++}, consists of two modules, \textit{Feature Extraction} and \textit{Classification}.
        The feature extraction module learns a set of local descriptors based on sampled points and their neighboring points. It comprises three distinct types of blocks: \textit{sampling}, \textit{grouping}, and \textit{PointNet}. It learns a feature from an input point cloud. Once the features are extracted, the feature vector is subsequently mapped to classes using fully connected layers. 
        Let $\mathcal{P}=\left\{p_{i} \in \mathbb{R}^{3}\right\}_{i<N}$ represent a point cloud consisting of $N$ points, where $p_{i}$ denotes a point with coordinates $(x,y,z)$. The model is formulated as described in~\eqref{eq:ModelLD-PointNet++}:
        \begin{equation}
            f_i = \mathcal{A}_{j=1}^K \left( h_{\theta}(p_i, p_{ij}) \right)
            \label{eq:ModelLD-PointNet++}
        \end{equation}
        where $p_{i}$ corresponds to a sampled point from $\mathcal{P}$, $f_i$ is a feature vector representing the local region centered around the point $p_{i}$, $p_{ij}$ is a neighboring point of $p_{i}$ which encodes the local information of $p_{i}$, $h_{\theta}$ represents a non-linear mapping function with learnable parameters denoted by $\theta$, and $\mathcal{A}$ refers to a symmetric aggregation operation over the $K$ neighboring points of $p_i$ (\textit{MAX} function, in our case).
        In~\eqref{eq:ModelLD-PointNet++}, three parts are important: \textbf{(1)}~\textit{Sampled point ($p_{i}$)}: These points serve as representative points of the local region. \textbf{(2)}~\textit{Neighborhood of a sampled point ($p_{ij}$)}: This corresponds to the function provided by the second block of the feature extraction module. \textbf{(3)}~\textit{Feature learning operation~($h_{\theta}$ and $\mathcal{A}$)}: These operations are carried out using PointNet, which constitutes the third block of the feature extraction module.
        
        \textit{Local Region Selection.}
        To describe the local region around a point $p_{i}$, we need to describe how to select the neighboring points of $p_{i}$. This task involves defining and parameterizing the neighborhood. The appropriate choice of a neighborhood definition depends on the characteristics of the point cloud. In the existing literature, two neighborhood definitions are commonly used~\cite{9127813}: Spherical neighborhood and KNN. The spherical neighborhood is defined by the radius and also a maximum number of neighbors limit. Unlike kNN, this type of neighborhood maintains the local region fixed, making the features more generalizable in space. This is advantageous for tasks that depend on recognizing local patterns~\cite{PointNet++}. Therefore, we adopt the definition of the spherical neighborhood as a versatile definition. The parameters of the neighborhood depend on the structure and density of the points in the data. The choice of these parameters is done empirically.
        
        \textit{PointNet for Global-Local Feature Extraction.}
        To extract features from the local region, we employ a PointNet~\cite{PointNet}. As illustrated in Fig.~\ref{fig:LD-PointNet++}, this block encodes a local region into a feature vector by taking $M$ local regions of neighboring points as input and generating $M$ feature vectors.
        In summary, after dividing the point cloud into $M$ local regions centered around a sampled point, we encode the neighbors of the sampled point using PointNet into a feature vector. Once PointNet is applied to the local regions, we have a set of feature vectors representing all the local regions. We apply PointNet again at this stage to this group of vectors to finally obtain a learnable representation of the point cloud.
        
    \subsection{Large Neighborhoods $\to$ Significant Improvements: Maximizing Low-Density 3D Point Cloud Classification}
    In computer vision, the receptive field is crucial for developing neural networks and enhancing their performance. Originally applied with success in 2D vision, it refers to the region of input data that influences the output of a neural unit, thereby marking the "visible" area of the image for that unit. Increasing the size of the receptive field significantly enhances the network's ability to capture features. Extending this concept to the 3D vision increases the capability of neural networks to interpret complex spatial structures, which is essential for understanding and extracting more informative features from 3D point cloud.
    In 2D, techniques such as the use of deeper hierarchical architectures or increasing convolution kernel sizes are employed to extend the receptive field. Transitioning these techniques from 2D vision to 3D point cloud analysis requires adapting these strategies, considering the unique properties of 3D point clouds such as their irregularity and density variability, which can impact the local features.
    
    To effectively increase the receptive field in neural networks dealing with 3D point clouds, two main strategies can be considered, analogous to those in 2D: \textbf{(1)} the use of hierarchical architectures, and \textbf{(2)} expanding the neighborhood (the local region) of each point by increasing the radius in the case of spherical neighborhoods. While hierarchical structures were adopted in the original PointNet++, our LD-PointNet++ focuses on neighborhood expansion, as illustrated in Table~\ref{tab:pointnet_comparison}. This approach improves neural network performance, allowing for the consideration of a broader context, crucial for accurate classification in low-density areas.
    The main modification to consider the context of low-density point clouds is the expansion of the spherical neighborhood. Indeed, the classification model for low-density point cloud can learn incomplete representations of the underlying structures due to the lack of points in the point cloud. By enlarging the neighborhood, more neighboring points are included, increasing the number of points considered and enabling a more comprehensive understanding of the overall structure of the 3D object/scene and local context. Consequently, the model can make more informed decisions, thereby enhancing performance.
    \begin{table}[tbp]
        \centering
        \caption{Comparison of PointNet++ and LD-PointNet++ for low-density 3D point cloud classification.}
        \label{tab:pointnet_comparison}
        \sisetup{table-format=-1.2}
        \resizebox{\columnwidth}{!}{
        \begin{tabular}{>{\bfseries}lcc}
            \toprule
            Property & \textbf{PointNet++} & \textbf{LD-PointNet++} \\
            \midrule
            Enlarged Neighborhood & \xmark & \cmark \\
            Hierarchical Architecture & \cmark & \xmark \\
            Technique to Scale & \multirow{2}{*}{Hierarchical architecture} & \multirow{2}{*}{Larger neighborhood} \\
            the Receptive Field &  &  \\

            Designed for Low-Density & \xmark & \cmark \\
            \bottomrule
        \end{tabular}
        }
    \end{table}   

\begin{figure}[bp]
        \centering
        \includegraphics[width=0.95\columnwidth]{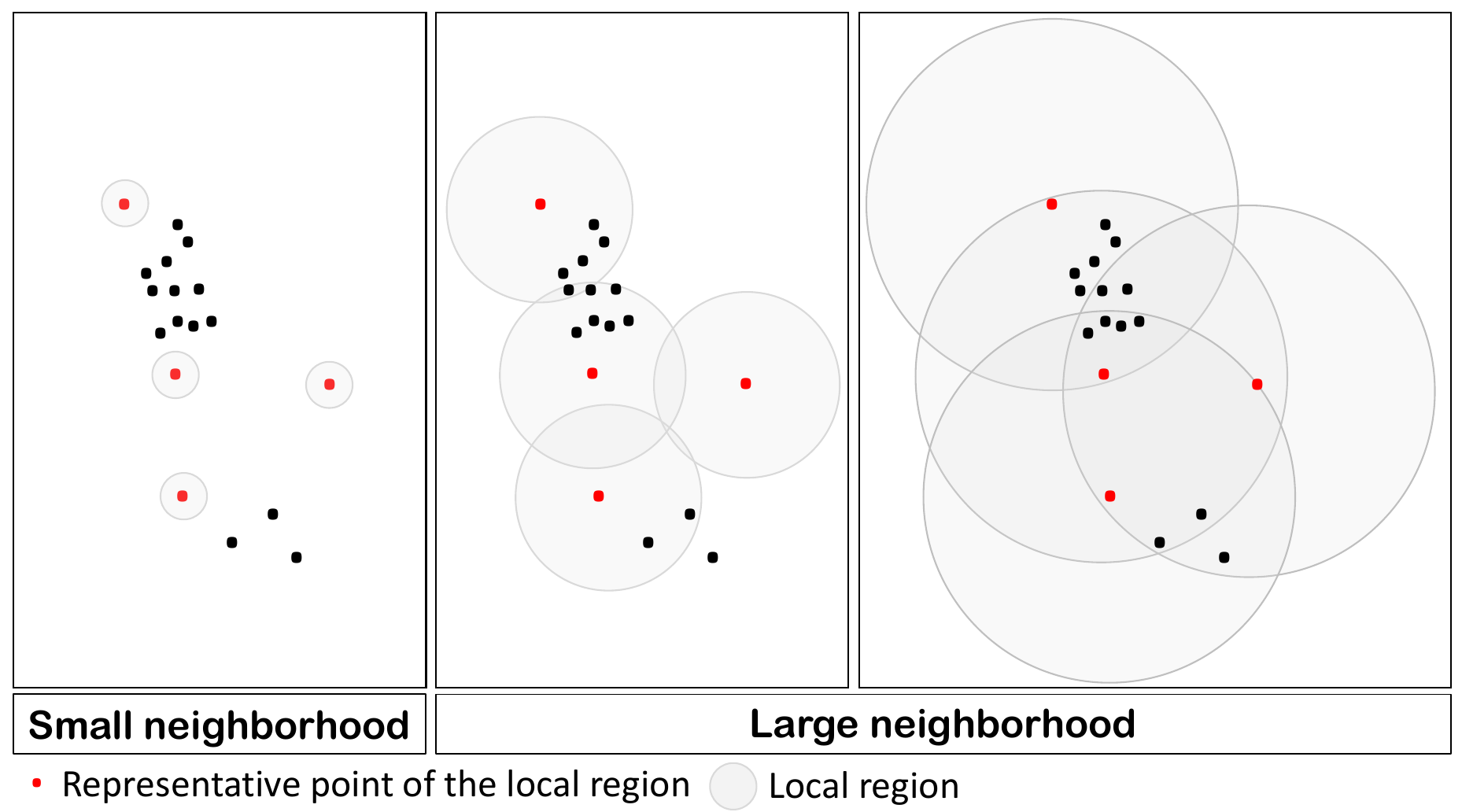}
        \caption{Large neighborhood effect in low-density 3D point cloud classification}
        \label{fig:NeighborhoodEffect}
    \end{figure}

    In regions with sparse points, the neighborhood encompasses a broader area, ensuring sufficient points are considered for classification. Conversely, in denser regions, the model can focus on a smaller area for classification (see Fig.~\ref{fig:NeighborhoodEffect}).
    
\begin{figure}[tbp]
        \centering
        \includegraphics[width=0.8\columnwidth]{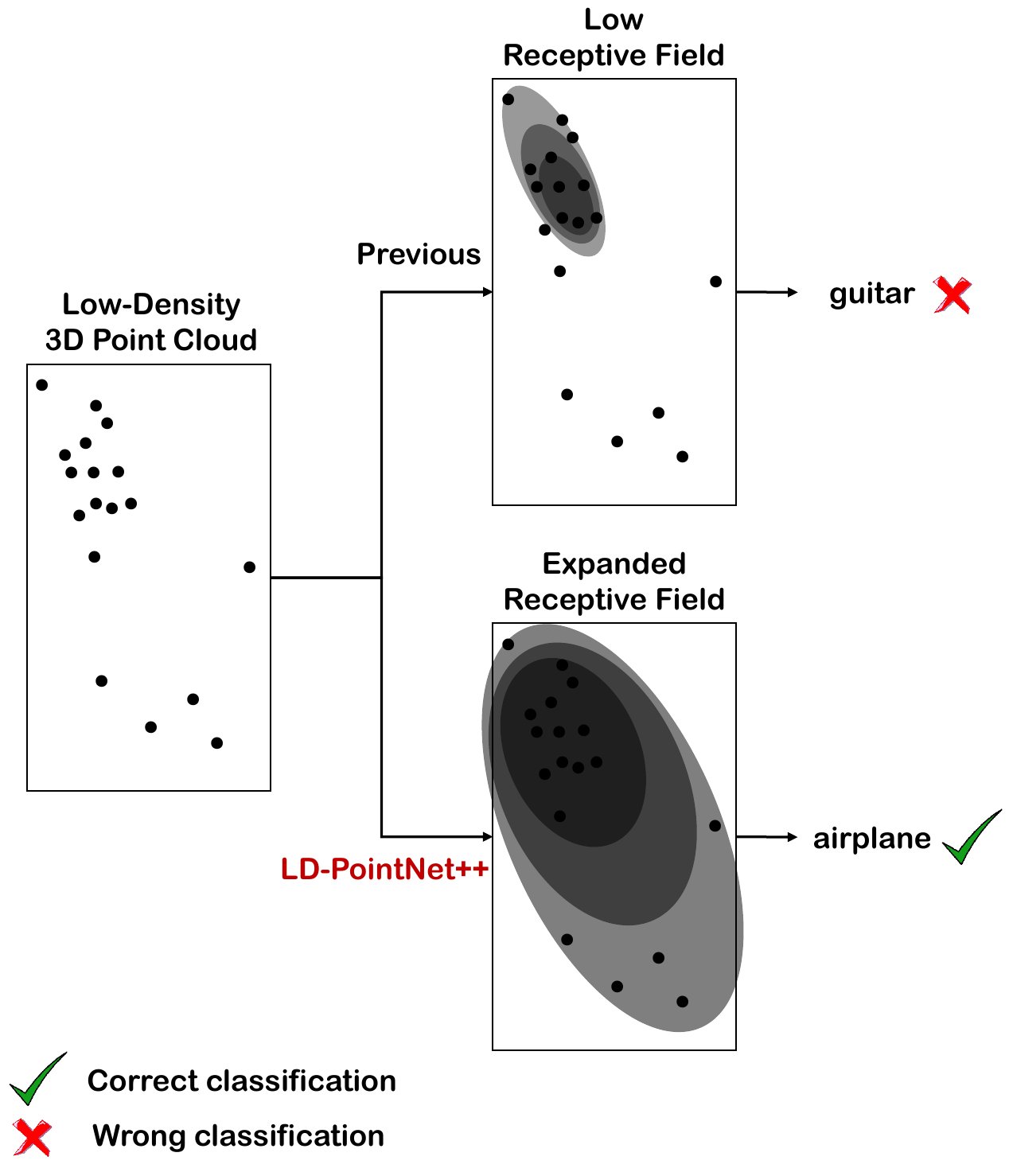}
        \caption{LD-PointNet++ expands receptive field in low-density 3D point cloud, improving feature capture and 3D structure recognition.}
        \label{fig:ReceptiveField}
    \end{figure}

    The use of a large neighborhood for low-density point cloud classification offers the advantage of naturally expanding the receptive field of neural networks. As a result, each neuron's receptive field extends over a wider area of the input point cloud, including a larger number of neighboring points, which allows capturing more extensive features, both global and local, and better understanding the overall structure of the 3D object (see Fig.~\ref{fig:ReceptiveField}).
    Formally, the volume $V(R)$ of a spherical neighborhood with radius $R$ around a point $p$ in a point cloud $\mathcal{P}$ is defined by the formula for the volume of a sphere:
    \begin{equation}
      V(R) = \frac{4}{3} \pi R^3  
      \label{eq:neighborhood}
    \end{equation}
    Thus, increasing $R$ results in a cubic increase in the volume~$V(R)$, which geometrically expands the area that each neuron can access and analyze.
    The impact on the receptive field can be expressed by~\eqref{eq:neighborhood}. This directly broadens the network’s capacity to analyze and understand the global structure of the 3D object/scene, facilitating a more precise classification, especially in low-density regions. In practice, this means that each neuron has access to a larger portion of the point cloud, enabling it to capture more extensive features, both global and local.
    Moreover, by increasing the radius $R$, the probability that a point $p$ is isolated, i.e., without neighbors in his neighborhood, significantly decreases. This is particularly crucial in low-density regions. By reducing the isolation of points, the neural network can better grasp the underlying structures, even in sparse regions, thereby facilitating a more precise and robust classification of the point cloud.
    
    The choice of neighborhood radius is critical and must be adapted to the specific characteristics of each dataset. An optimal radius must be large enough to capture the necessary contextual data but not so vast that it includes irrelevant points that could dilute the significance of local structures. Achieving this balance requires empirical testing and adjustments based on the specific characteristics of each dataset.

%% file: sections/04_experimental_evaluation.tex
\section{Experimental Evaluation of LD-PointNet++}
\label{section:Experiments}
    \subsection{Experimental Setup}
    We conducted our experiments using the ModelNet dataset~\cite{ModelNet} to train and test our model. During the training phase, we use a fixed set of 1024 points per object, whereas for testing, we reduced the points to 64 per object, as discussed in Section~\ref{section:RobustnessAnalysis}. The ModelNet dataset consists of two variants: ModelNet40, detailed in Section~\ref{section:RobustnessAnalysis}, and ModelNet10, containing 4899 CAD models from 10 distinct object categories. We adhered to the standard data split procedure~\cite{PointNet++}. 
    The training process involved training the network from scratch using the Adam optimizer on an Nvidia GeForce RTX~3090. We trained the model over 200 epochs with a batch size of 24. The initial learning rate was 0.001, and we implemented a decay strategy, reducing the learning rate by~0.0001. Furthermore, we used the cross-entropy loss function.
    We employed a spherical neighborhood with a radius of 0.8, allowing for a maximum of 64 neighbor points within each local region. Subsequently, each PointNet in the architecture consists of three fully connected layers, with the first PointNet having widths [128, 128, 256], and the second having widths~[256, 512, 1024]. We used the same classification module as~\cite{PointNet++}.
    
    \subsection{Performance Evaluation in Low-Density Context}
    This section is dedicated to the evaluation of the classification process conducted by LD-PointNet++ in the context of low-density scenarios. In order to gain a comprehensive understanding of the performance, both quantitative and qualitative assessments were employed.
    
\begin{figure}[!h]
            \centering
            \includegraphics[width=0.8\columnwidth]{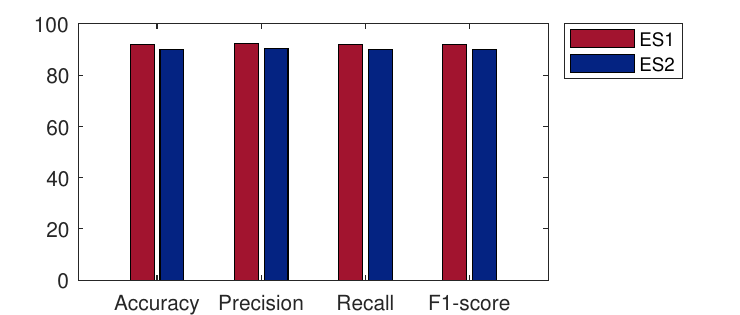}
            \caption{Classification results of LD-PointNet++ on the ModelNet40 dataset under experimental setups ES1 and ES2.}
            \label{fig:precision_recall_LDPointNet++}
        \end{figure}

        \parasection{Quantitative Evaluation.}
        To quantitatively evaluate the performance of LD-PointNet++, we compare the different metrics, precision, recall, and F1-score, under two experimental setups~(ES1 and ES2): one where the model is tested with 1024 points and another with 64 points, as depicted in Fig.~\ref{fig:precision_recall_LDPointNet++}. 
        
\begin{figure}[!h]
        \centering
          \subfloat[ES 1]{%
            \includegraphics[width=0.485\columnwidth]{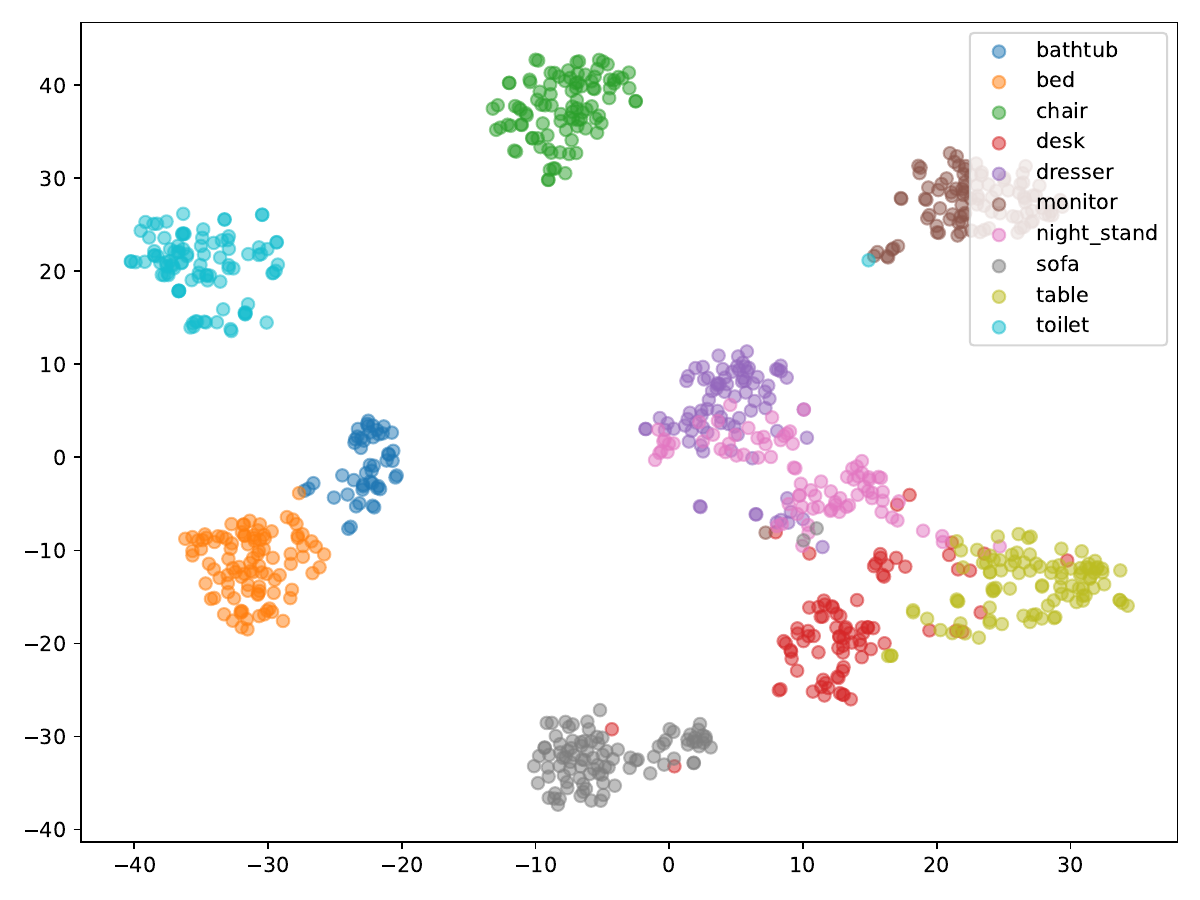}
            \label{fig:TSNELD-FE-ES1}
            }\hfill
          \subfloat[ES 2]{%
            \includegraphics[width=0.485\columnwidth]{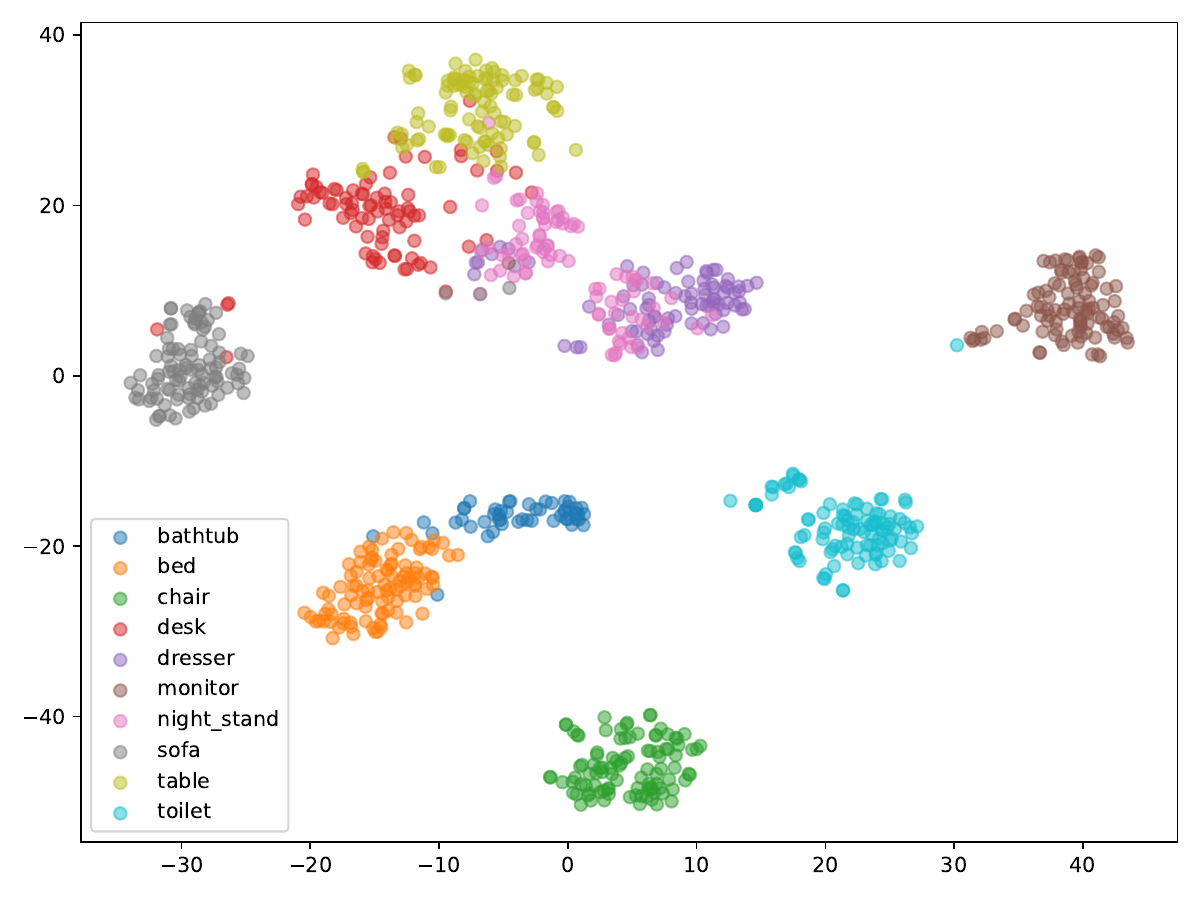}
            \label{fig:TSNELD-FE-ES2}
            }\\
          \caption{T-SNE visualization of the features learned by LD-PointNet++ under experimental setups ES1 and ES2.}
          \label{fig:tsne_LD-PointNet++}
        \end{figure}

        \parasection{Qualitative Evaluation.}
         To demonstrate the effectiveness of the learned representation of point cloud, we employed t-SNE visualization to assess the separability of the features. We applied t-SNE to the learned features in the two experimental setups, ES1 and ES2. The results are summarized in Fig.~\ref{fig:tsne_LD-PointNet++}.
        
    \subsection{Comparison with State-of-the-Art}
    In this section, we present a comparative analysis of our proposed LD-PointNet++ model against existing state-of-the-art methods for low-density 3D point cloud classification. As per standard practice, we present in Table~\ref{tab:OA-LDPointNet++} the overall accuracy (OA) and the mean class accuracy (mAcc) based on a test set where each point cloud consists of 64 points.
        \begin{table}[tbp]
        \centering
        \caption{Comparison of LD-PointNet++ with state-of-the-art for low-density 3D point cloud classification on ModelNet40 dataset (Training with 1024 points, Testing with 64 points).}
        \sisetup{table-format=-1.2}
        \label{tab:OA-LDPointNet++}
        \resizebox{0.88\columnwidth}{!}{
        \begin{tabular}{lrr}
        \toprule 
        \textbf{Method}& \textbf{OA} & \textbf{mAcc} \\ \midrule
        PointNet++~\cite{PointNet++}  &35.39 \% & 37.35 \% \\ 
        Point-TnT~\cite{Point-TnT}  & 38.16 \% & 34.39 \% \\ 
        DGCNN~\cite{dgcnn}  & 4.61 \% & 5.05 \% \\ 
        PointMLP~\cite{PointMLP} &4.29 \% & 7.75 \% \\ 
        \rowcolor{LightGray} \textbf{LD-PointNet++}~(Ours) & \textbf{90.10 \%} & \textbf{~~~ 86.43 \%} \\ \bottomrule
        \end{tabular}
        }
        \end{table}
        
    \subsection{Discussion}
    \balance

\parasection{How effective is LD-PointNet++ in low-density 3D point cloud classification?}
The results presented in the previous sections demonstrate the effectiveness of LD-PointNet++ in low-density 3D point cloud classification. Although it does not achieve the highest accuracy in dense point cloud scenarios~(see Fig.~\ref{fig:RobustesseSampling}), LD-PointNet++ shows very promising results in low-density scenarios. The success of our method is due to the use of a large neighborhood, which naturally expands the receptive field and efficiently captures both global and local features.
LD-PointNet++'s robustness in the low-density scenario is confirmed by analyzing classification metrics as shown in Fig.~\ref{fig:precision_recall_LDPointNet++}. Even with a reduced number of input points (ES2), LD-PointNet++ achieves an F1-score of~90.16\%, only 1.96\% less than when using the full set of input points (ES1). This demonstrates the consistent and reliable performance of LD-PointNet++ in low-density 3D point cloud classification.
Additionally, we visualize the learned features using t-SNE to show feature separability. As depicted in Fig.~\ref{fig:tsne_LD-PointNet++}, even for the low-density scenario~(ES2), LD-PointNet++ maintains good separation between classes, supporting the quantitative results. 

\parasection{How does LD-PointNet++ compare to state-of-the-art methods in low-density scenarios?}
As shown in Table~\ref{tab:OA-LDPointNet++}, LD-PointNet++ clearly outperforms all other methods in the low-density classification scenario, especially in terms of overall accuracy (OA) for the same 64 point input: an improvement of 54.71\% compared to PointNet++ and up to 85.81\% compared to PointMLP. This highlights LD-PointNet++'s ability to excel in very low-density point cloud scenarios.
    
\parasection{How robust is LD-PointNet++ to variations in input density?}
The robustness of LD-PointNet++ to input density variations is also studied, as shown in Fig.~\ref{fig:RobustesseSampling}. LD-PointNet++ outperforms other methods across various input density ratios, particularly in low-density scenarios, with a precision margin of up to 85.81\% compared to PointMLP. In addition, LD-PointNet++ achieves almost constant accuracy, exceeding 90\%, for all sampling ratios. To the best of our knowledge, there is no previous method that can achieve the same level of classification accuracy with such a low level of input point density.

%% file: sections/05_conclusion.tex
\section{Conclusion}
\label{section:Conclusion}
In this paper, we introduced LD-PointNet++, a novel variant specifically designed for low-density 3D point cloud classification, featuring a larger neighborhood. Through a robustness study, we evaluated the performance of existing 3D point cloud classification methods, highlighting their limitations in low-density contexts. 
The experiments conducted on the ModelNet dataset demonstrated that LD-PointNet++ outperforms existing methods, consistently achieving a classification accuracy of over 90\%, regardless of point cloud density. For instance, even with a reduced subset of 64 points, LD-PointNet++ achieved an accuracy of 90.1\%. While LD-PointNet++ may not reach the highest accuracy in dense point cloud scenarios, it stands out when evaluated with a reduced number of points.

%% file: references.bib
@ARTICLE{9127813,
  author={Guo, Yulan and Wang, Hanyun and Hu, Qingyong and Liu, Hao and Liu, Li and Bennamoun, Mohammed},
  journal={IEEE Transactions on Pattern Analysis and Machine Intelligence}, 
  title={Deep Learning for 3D Point Clouds: A Survey}, 
  year={2021},
  volume={43},
  number={12},
  pages={4338-4364},
  doi={10.1109/TPAMI.2020.3005434}}

@ARTICLE{9857670,
  author={Fei, Ben and Yang, Weidong and Chen, Wen-Ming and Li, Zhijun and Li, Yikang and Ma, Tao and Hu, Xing and Ma, Lipeng},
  journal={IEEE Transactions on Intelligent Transportation Systems}, 
  title={Comprehensive Review of Deep Learning-Based 3D Point Cloud Completion Processing and Analysis}, 
  year={2022},
  volume={23},
  number={12},
  pages={22862-22883},
  doi={10.1109/TITS.2022.3195555}}

@INPROCEEDINGS{PointNet,
  author={Charles, R. Qi and Su, Hao and Kaichun, Mo and Guibas, Leonidas J.},
  booktitle={2017 IEEE Conference on Computer Vision and Pattern Recognition (CVPR)}, 
  title={Point{N}et: Deep Learning on Point Sets for 3D Classification and Segmentation}, 
  year={2017},
  volume={},
  number={},
  pages={77-85},
  doi={10.1109/CVPR.2017.16}}

@INPROCEEDINGS{PointNet++,
  author = {Qi, Charles R. and Yi, Li and Su, Hao and Guibas, Leonidas J.},
  title = {PointNet++: Deep hierarchical feature learning on point sets in a metric space},
  year = {2017},
  booktitle = {Advances in Neural Information Processing Systems},
  volume = {2017-December},
  pages = {5100 – 5109},
  doi={10.5555/3295222.3295263}}

@INPROCEEDINGS{PointNeXt,
 author = {Qian, Guocheng and Li, Yuchen and Peng, Houwen and Mai, Jinjie and Hammoud, Hasan and Elhoseiny, Mohamed and Ghanem, Bernard},
 booktitle = {Advances in Neural Information Processing Systems},
 pages = {23192--23204},
 title = {Point{N}e{X}t: Revisiting PointNet++ with Improved Training and Scaling Strategies},
 volume = {35},
 year = {2022}
}

@INPROCEEDINGS{PointMLP,
title={Rethinking Network Design and Local Geometry in Point Cloud: A Simple Residual {MLP} Framework},
author={Xu Ma and Can Qin and Haoxuan You and Haoxi Ran and Yun Fu},
booktitle={International Conference on Learning Representations},
year={2022}
}

@INPROCEEDINGS{Pointwise,
  author={Hua, Binh-Son and Tran, Minh-Khoi and Yeung, Sai-Kit},
  booktitle={2018 IEEE/CVF Conference on Computer Vision and Pattern Recognition}, 
  title={Pointwise Convolutional Neural Networks}, 
  year={2018},
  volume={},
  number={},
  pages={984-993},
  doi={10.1109/CVPR.2018.00109}}

@INPROCEEDINGS{KPConv,
  author={Thomas, Hugues and Qi, Charles R. and Deschaud, Jean-Emmanuel and Marcotegui, Beatriz and Goulette, François and Guibas, Leonidas},
  booktitle={2019 IEEE/CVF International Conference on Computer Vision (ICCV)}, 
  title={{KPC}onv: Flexible and Deformable Convolution for Point Clouds}, 
  year={2019},
  volume={},
  number={},
  pages={6410-6419},
  doi={10.1109/ICCV.2019.00651}}

@INPROCEEDINGS{Zhang2020AECNN,
  author={Zhang, Junming and Yu, Ming-Yuan and Vasudevan, Ram and Johnson-Roberson, Matthew},
  booktitle={2020 International Conference on 3D Vision (3DV)}, 
  title={Learning Rotation-Invariant Representations of Point Clouds Using Aligned Edge Convolutional Neural Networks}, 
  year={2020},
  volume={},
  number={},
  pages={200-209},
  doi={10.1109/3DV50981.2020.00030}}

@ARTICLE{Zhang2022RIConv++,
  title={RIConv++: Effective Rotation Invariant Convolutions for 3D Point Clouds Deep Learning},
  author={Zhang, Zhiyuan and Hua, Binh-Son and Yeung, Sai-Kit},
  journal={International Journal of Computer Vision},
  volume={130},
  number={5},
  pages={1228--1243},
  year={2022},
  doi={10.1007/s11263-022-01601-z}}

@ARTICLE{dgcnn,
  title={Dynamic Graph CNN for Learning on Point Clouds},
  author={Wang, Yue and Sun, Yongbin and Liu, Ziwei and Sarma, Sanjay E. and Bronstein, Michael M. and Solomon, Justin M.},
  journal={ACM Transactions on Graphics (TOG)},
  year={2019},
  doi={10.1145/3326362}}

@INPROCEEDINGS{Rgcnn,
  title={{RGCNN}: Regularized Graph {CNN} for Point Cloud Segmentation},
  author={Gusi Te and Wei Hu and Zongming Guo and Amin Zheng},
  booktitle={ACM international conference on Multimedia},
  year={2018},
  doi={10.1145/3240508.3240621}}

@INPROCEEDINGS{Point-TnT,
  author={Berg, Axel and Oskarsson, Magnus and O’Connor, Mark},
  booktitle={2022 26th International Conference on Pattern Recognition (ICPR)}, 
  title={Points to Patches: Enabling the Use of Self-Attention for 3D Shape Recognition}, 
  year={2022},
  volume={},
  number={},
  pages={528-534},
  doi={10.1109/ICPR56361.2022.9956172}}

@INPROCEEDINGS{ModelNet,
  author={Zhirong Wu and Song, Shuran and Khosla, Aditya and Fisher Yu and Linguang Zhang and Xiaoou Tang and Xiao, Jianxiong},
  booktitle={2015 IEEE Conference on Computer Vision and Pattern Recognition (CVPR)}, 
  title={3D ShapeNets: A deep representation for volumetric shapes}, 
  year={2015},
  volume={},
  number={},
  pages={1912-1920},
  doi={10.1109/CVPR.2015.7298801}}

@ARTICLE{t-SNE,
  author  = {Laurens van der Maaten and Geoffrey Hinton},
  title   = {Visualizing Data using t-{SNE}},
  journal = {Journal of Machine Learning Research},
  year    = {2008},
  volume  = {9},
  number  = {86},
  pages   = {2579--2605}
}
